# Object Recognition
# with Imperfect Perception and Redundant Description

**Claude Barrouil - Jérôme Lemaire\***
Centre d'Études et de Recherches de Toulouse
Département d'Études et de Recherches en Automatique,
2 avenue Belin, B.P. 4025, 31055 Toulouse cedex, FRANCE.
E-mail: {barrouil,lemaire}@cert.fr

## Abstract

This paper deals with a scene recognition system in a robotics context. The general problem is to match images with *a priori* descriptions. A typical mission would consist in identifying an object in an installation with a vision system situated at the end of a manipulator and with a human operator provided description, formulated in a pseudo-natural language, and possibly redundant. The originality of this work comes from the nature of the description, from the special attention given to the management of imprecision and uncertainty in the interpretation process and from the way to assess the description redundancy so as to reinforce the overall matching likelihood.

## Introduction

In a robotics context, a scene interpretation system achieves the transformation of visual images into semantic descriptions of the world that can interface with other decision processes and elicit appropriate actions [CK92]. In the project at hand, the scene recognition problem is defined as the matching of images with *a priori* descriptions; the descriptions are given in a high level language close to natural language and without numerical data. Very few studies seem to exist on this type of problem. The only related work is Dubois & Jaulent's which expands fuzzy region labeling methods for performing the recognition of objects described by a human operator in a context of bidimensionnal scenes, in synthesis images without object coverings [Jau86].

The originality of this work comes from the nature of the description, from the special attention given to the management of imprecision and uncertainty in the interpretation process and from the way to assess the description redundancy so as to reinforce the overall matching likelihood. This paper is organized as follows: in section 1, the goal of the project and a discussion on the modeling of uncertainty and imprecision in computer vision are presented ; in section 2, a method for identifying an object with uncertain data and a priori given description is explained; in section 3, results of the first version of the system are given and it is emphasized that, generally, the matching quality decreases as the description is more detailed; then, in section 4, a method is proposed to define redundancy, and to use it so as to reinforce the matching quality.

## 1   Position of the problem

### 1.1   Goal of the project

The applicative scenario consists in a mobile robot with a color ccd-camera at the end of an on-board manipulator. The environment is a complex compressed air installation that occupies a significant volume, including floodgates on different pipes cranked and interconnected with T-squares. The pipes join different objects: a cistern, a super-charger... The mission goal is to identify the position of a given floodgate. The environment is only known through the fuzzy hints of the mission description, and the floodgate position is generally given by means of a path that leads to it. Examples of floodgates designations are given below for a better understanding:

• **cistern** *above* **pipe** *connected_to* **elbow** *connected_on_the_right_to* **pipe** *on* **floodgate[hunt]**
• **super-charger** *in_front_of* **horizontal pipe** *connected_on_the_left_to* **elbow** *connected_to* **pipe** *on* **red floodgate[hunt]**

A scene interpretation system in charge of "scene recognition" has to perform the following aspects: Description of the Expected Scene (DES), Description of the Perceived Scene (DPS - the perceived entities & their attributes), interpretation of the scene as the matching of both descriptions [Lem95, LLB96]. Matching is defined as the comparison of two representations in order to discover their similarities and their differences. This mechanism transforms the given representations into a more abstract one. In this part, it is necessary to represent the objects matching confidence, which are obtained by aggregating local matching confidence for types, attributes, and spatial relationships. The mission description data is in the DES and includes a path that should lead to the hunt object. The DES must be compared to the sensor provided data. A single shot is generally not sufficient to match the DES. The system must build a environment map that integrates different DPS. Then, the map is compared with the DES and the analogies recorded in the DRS (Description of the Recognized Scene).

---
*must regain : GESMA, BP42, 29240 Brest Naval - FRANCE.

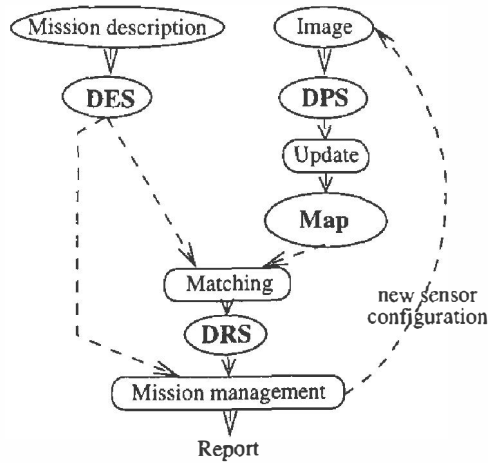

Figure 1: **System overview.**

### 1.2  Uncertainty and Scene recognition

Uncertainty abounds in every phase of computer vision. In a scene interpretation system, the knowledge is often uncertain and/or imprecise. On the one hand, the percepts extracted from images are generally imprecise and the combination of percepts in order to identify objects in the scene also introduces uncertainty. On the other hand, knowledge bases about the environment contain imprecision and uncertainty, and the description of an expected scene given in a pseudo-natural language is evidently vague. Imprecision refers to the fact that the value of the variable under consideration is only known to belong to a subset of values which is not a singleton. Uncertainty pertains to the lack of complete information which forbids to consider as certain the belonging or non-belonging of the variable to a given subset. It is also important to awake to the importance of the modeling and the treatment of uncertainty and imprecision throughout the interpretation process as regards in the performance of the system. Different mathematical frameworks are available for modeling uncertainty, such as the probability theory, Dempster-Shafer's belief functions, or the possibility theory.

A probability measure represents an experimental observation of the realization frequency of an event or a subjective knowledge assigned to an event. The use of probabilities is often bound with the Bayes' rule. The assumptions on which this approach is based are not often verified. It requires the mutual exclusiveness and exhaustiveness of the hypotheses, and the independence of the different events over against one hypothesis. This method requires also a large amount of data to determine the estimates for the prior and conditional probabilities. For the interpretation stage, this approach seems too rigid. But, in the field of image segmentation and 3D reconstruction, good results have been obtained in various computer vision applications.

The belief theory proposed by Shafer was developed within the framework of Dempster's work on upper and lower probabilities induced by a multi-valued mapping. This theory introduces the notion of mass of probability and an explicit modeling of ignorance. Two problems are generally emphasized with this approach. The first one stems from the computational complexity. The second one results from the normalization process which can lead to incorrect and counter-intuitive results. In computer vision, experiments have been realized for the reasoning on the events in the scene interpretation system VISIONS [DCB89].

The possibility theory and fuzzy sets [DP88] offer a framework which allows pieces of information which are both imprecise and uncertain to be modeled. The main features are:
- the faithfulness of the representation of subjective data,
- no need for a *priori* knowledge,
- a convenient and straightforward formalism to express and aggregate data with different modes of combination.

This last aspect is very important because the problem of the independence of sources in the combination of uncertainty and imprecision often appears in scene interpretation. The aspect of the texture of an object and its geometry are not really independent. On the contrary, the type of an object is generally independent of its color. Therefore, the possibility theory is attractive in that context because first, no unique combination mode is imposed and second, the choice of the combination mode depends on an assumption about the reliability of sources and on the nature of the information. In computer vision, this approach has gained popularity in applications on perception fusion or on scene interpretation [SG94].

## 2  Scene recognition in presence of uncertainty

### 2.1  Data uncertainty

#### 2.1.1  Description of the Expected Scene

**Definition :** *The Description of the Expected Scene (DES) is the description of the position of the hunt*

*object by means of a path that leads to it through the installation.*

The DES is composed of two sets. Let $\mathcal{O} = \{O_i\}$ be the set of the expected objects with their features, and $\mathcal{R} = \{\mathcal{R}_k\}$ with $\mathcal{R}_k = \mathcal{R}_k(O_i...O_{i'})$ being the set of the relations between the objects. A logic framework has been chosen for the syntax of the expressions, because the conjunction, disjunction and negation are necessary in order to express the complexity of the installation and the uncertainties induced by the operator. A pseudo-natural language has been defined for the implementation of the DES (see [Lem95, LLB96] for more details).

The description of an object of the DES gives the necessary conditions for the possible matching of a perceived object with an expected object. If a perceived object $\Omega_i$ can correspond to the expected object $object_1$ that is described for example as a *blue and vertical pipe*, it must have the properties :

$correspond\_to(object_1, \Omega_i)$
$\rightarrow pipe(\Omega_i) \wedge blue(\Omega_i) \wedge vertical(\Omega_i).$

Other descriptions use the disjunction and / or the negation because the operator can have doubts about the properties of the expected object :

$correspond\_to(object_2, \Omega_i)$
$\rightarrow floodgate(\Omega_i) \wedge (red(\Omega_i) \vee blue(\Omega_i))$

$correspond\_to(object_3, \Omega_i)$
$\rightarrow floodgate(\Omega_i) \wedge (\neg red(\Omega_i))$

The relations between the objects are geometric or topological. The different relations are combined in order to obtain a composed relation or an alternative in the definition of the relation. The relations are described by necessary and sufficient conditions. The relation $relation_i$ (the $i_{th.}$) of the DES that concerns $n$ objects ($relation_i \in \mathcal{L}_{R_n}$) is formulated by : $relation_i(O_1, O_2, ..., O_n) \leftrightarrow$ conjunctions and disjunctions of relations of $\mathcal{L}_{R_n}$ applied to the objects $O_1, ..., O_n$. The relation that is in the description $O_1 - above$ & $on\_the\_right\_to - O_2$ is formulated by : $relation_1(O_1, O_2) \leftrightarrow above(O_1, O_2) \wedge on\_the\_right\_to(O_1, O_2)$.

The disjunction can also be in the definition of the relation because it is possible that the operator should not remember the exact configuration of the installation. For example :

$relation_j(O_1, O_2)$
$\leftrightarrow (in\_front\_of(O_1, O_2)) \vee on\_the\_left\_to(O_1, O_2),$

$relation_k(O_1, O_2)$
$\leftrightarrow near\_from(O_1, O_2) \wedge (\neg connected\_to(O_1, O_2)).$

In order to express the path with more freedom, other possibilities have to be offered. If the operator has doubts, he may want to define more than one path to lead to the hunt object. The installation is perhaps well known by the operator but he hesitates between two positions for the hunt object. In those cases, the object has to be defined by a disjunction in the localization : $localization(object_{r.}) =$
$relation(object_{r.}, object_i) \vee relation(object_{r.}, object_j)$

Another possibility consists in the definition by the operator of an and / or graph because he hesitates between two (or more) graphs for the DES.

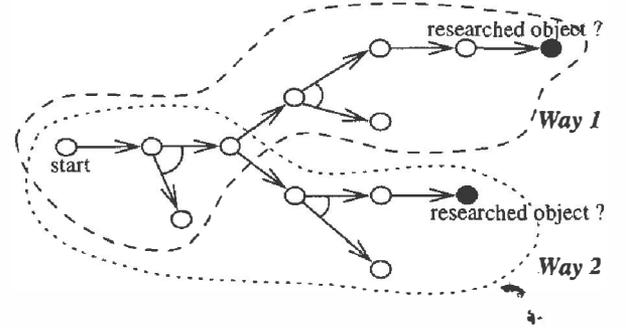

Figure 2: **Graph with 2 possible ways.**

### 2.1.2   Description of the Perceived Scene

**Definition :** *The Description of the Perceived Scene (DPS) is the set of data that come from the analysis of the information given by the sensor from a given position. The perceived data are expressed thanks to the same language as for the expected objects.*

An image $\iota_t$ that corresponds to the position of the sensor $\kappa_t$ gives us a set of percepts (edges, regions) $p_t = \{p_{ti}\}$ that are put together by a first matching in order to constitute objects that are only defined by their types. We obtain the set $\omega_t = \{\omega_{ti}\}$ of the perceived objects in two dimensions. The following informations are computed : is it a pipe, a floodgate...? Numerical indications for positions and dimensions are saved by the system at this stage too.

The identification of objects is a two steps process: detection and validation. Detection is based on a segmentation step. The validation of the object hypotheses is based on the computing of a *global confidence degree* for each objects. This degree is a combination of several confidence degrees, each one concerning a particular attribute of the object (geometric attribute, aspect attribute...). The different attributes, the associated weights and the mode of combination depend on the object type ($\pi_{global} = G(\pi_1, \pi_2...\pi_k)$).

### 2.2   Scene recognition

In ordinary matching systems, the operations executed in the pattern matching procedure are based on the test of the identity of symbols appearing both in the pattern and in the data, or more generally, the belonging of attribute values of the data to sets prescribed by the pattern. It is necessary to have some flexibility in the pattern matching procedure, because data are pervaded with imprecision and uncertainty, and the pattern includes vague specifications. In this case, it is natural to introduce a valuation of the compatibility of the ill-known data with the tolerant pattern. Then a fuzzy matching procedure can be defined. First, we will define precisely the map and the DRS, and afterwards we will present the matching method with the uncertainty management.

### 2.2.1 The map and the DRS

**Definition :** *The map contains the data perceived from the different positions of the sensor. It is the set of the perceived objects that have not been eliminated through inconsistencies.*

The map is the set $\Omega = \{\Omega_i\}$ of the perceived 3D objects. They come from the fusion of the different objects $\omega_j$ found in the images. Every object $\Omega_i$ is a sub-set of the set of all the $\omega_{tj}$ (with $t \leq t_{present}$). It is important to underline the dynamic process for the constitution of the map.

**Definition :** *The Description of the Recognized Scene (DRS) is the part of the map that could correspond to the DES. In fact, it is the set of the possible matching between the DES and the map with a sufficient level of the confidence degree.*

The DSR contains the parts of the map that can be matched with the DES. It is composed of two sets :

- $\Theta = \{\Theta_i\}$, the set of the recognized objects. The attributes of these objects correspond to the attributes of certain objects of the DES (type, orientation...).

- $\Gamma = \{\Gamma_j\}$ where $\Gamma_j$ is a relation of arity $arity(\Gamma_j)$ that concerns objects $\Theta_i$ of $\Theta$ and that is verified when it can be a realization of the relation $\mathcal{R}$ between the objects $O_i$ matched with the $\Theta_i$.

### 2.2.2 The Matching procedure

From the available knowledge on the perceived objects $\{\Omega_j\}$ pertaining to their type, color, geometry... the system can compute the confidence degree $\pi(\Omega_j; O_i)$ that the perceived object $\Omega_j$ corresponds to the expected object $O_i$. For the description $O_i$ *is_a horizontal, blue or green, floodgate*, the possible matching of $\Omega_j$ with $O_i$ implies that floodgate($\Omega_j$) $\wedge$ horizontal($\Omega_j$) $\wedge$ ( blue($\Omega_j$) $\vee$ green($\Omega_j$) ) is verified with a sufficient confidence degree. The computation is performed within the framework of the possibility theory [DP88] (use of min / max) :
$$\pi(O_i(\Omega_j)) = $$
$$min\,(\pi\,(floodgate\,(\Omega_i))\,,\pi\,(horizontal\,(\Omega_i)))\,,$$
$$max\,(\pi\,(blue\,(\Omega_i))\,,\pi\,(green\,(\Omega_i)))\,.$$
It is interesting to notice that a precise identification of the possibility distributions is not required since they are not very sensitive to slight variations of the possibility degree. After this first step, the system gives the matching hypotheses independently of the spatial relations between the objects (Figure 3).

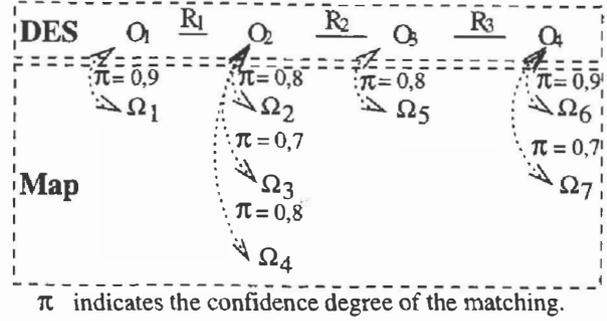

$\pi$ indicates the confidence degree of the matching.

Figure 3: **Results after the first step.**

At the second step, the system looks among the pairs of objects $(\Omega_j; \Omega_{j'})$ which could correspond (with a sufficient confidence degree), to a pair $(O_i; O_{i'})$ with $\pi(\gamma(\Omega_j\,;\,\Omega_{j'})\,,\,\mathcal{R}_n\,(O_i; O_{i'}))$ sufficiently large, where $\gamma_{jj'}$ denotes what is known of the spatial relation between $\Omega_j$ and $\Omega_{j'}$ and $\pi(\gamma_{jj'}; \mathcal{R}_{nii'})$ the compatibility of $\gamma_{jj'}$ with $\mathcal{R}_{nii'}$. Then, it is possible to compute the confidence degree $\pi(\Omega_j.\gamma.\Omega_{j'}; O_i.\mathcal{R}_n.O_{i'}) = F(\pi\,(\Omega_j; O_i)\,,\,\pi(\Omega_{j'}; O_{i'})\,,\,\pi(\gamma(\Omega_j; \Omega_{j'})\,,\,\mathcal{R}_n(O_i, O_{i'})))$ that the part of the "DES" $O_i.\mathcal{R}_n.O_{i'}$ could be seen in the scene with $\Omega_j$ and $\Omega_{j'}$. The aim is to go on with the recognition with all the expected objects and their relations that constitute the "DES". We first chose the classic $min$ operator for $F$.

The aim is to go on with the recognition with all the expected objects and their relations that constitute the "DES". At the beginning of the process, several pairs $(\Omega_j; \Omega_{j'})$ may correspond to a pair of expected objects $(O_i; O_{i'})$ with a sufficient confidence degree. Nevertheless, the matching process with other expected objects reduce the number of indecisions. An example of result

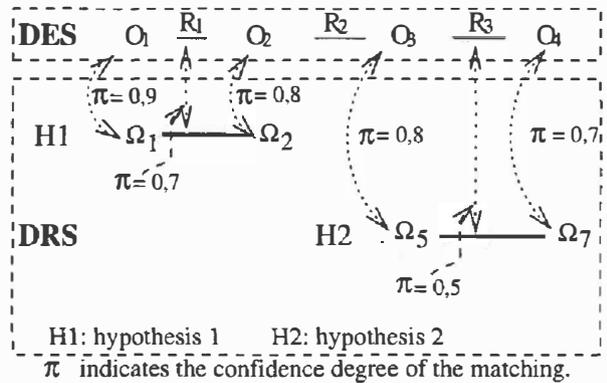

$\pi$ indicates the confidence degree of the matching.

Figure 4: **Example of result after the matching process.**

is given Figure 4 (the interpretation hypotheses with a sufficient confidence degree are only conserved). It is important to specify that the algorithm has to consider the possibility of objects being hidden or standing outside the shot.

## 3   Results and discussions

### 3.1   Results of the first version

All the concepts presented in the section 2 are implemented in the first version of the system. The experiments are based on a single shot taken on a modular PVC pipe installation in order to increase the complexity progressively. We give an example of Session with the System for the the data of Figure 5.

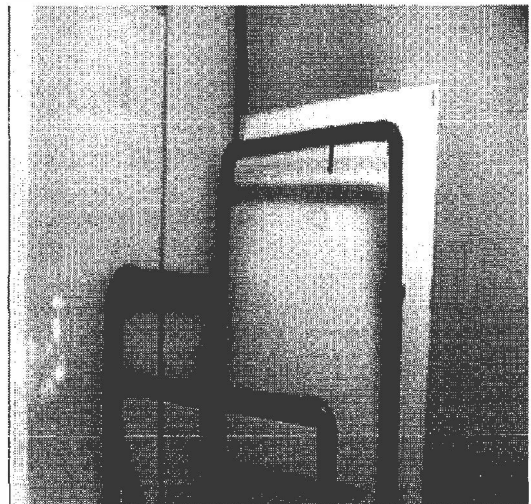
image of the installation

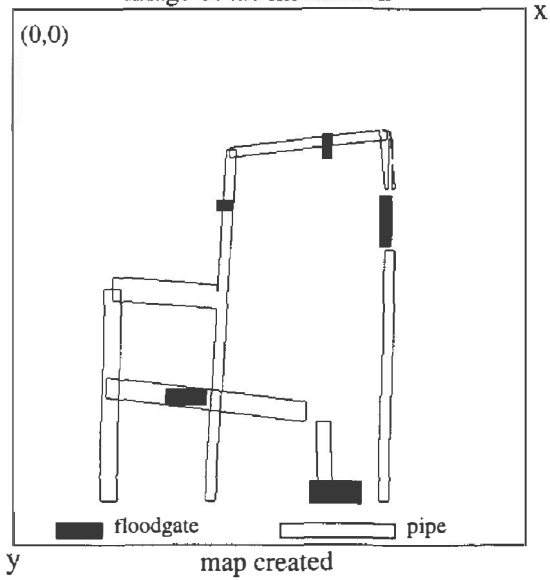
map created

Figure 5: **Results with a typical Scene.**

(Parameters of the perceived objects)

$\Omega_1$   Pipe (conf. degree= 0.37)
    $x \in [362; 367]$ , $y \in [116; 174]$
$\Omega_2$   Pipe (conf. degree= 0.88)
    $x \in [211; 364]$ , $y \in [120; 138]$
$\Omega_3$   Pipe (conf. degree= 0.97)
    $x \in [367; 368]$ , $y \in [124; 174]$
$\Omega_4$   Pipe (conf. degree= 0.81)
    $x \in [361; 367]$ , $y \in [235; 485]$
$\Omega_5$   Pipe (conf. degree= 0.76)
    $x \in [97; 199]$ , $y \in [274; 282]$
$\Omega_6$   Pipe (conf. degree= 0.58)
    $x \in [93; 97]$ , $y \in [274; 485]$
$\Omega_7$   Pipe (conf. degree= 0.74)
    $x \in [192; 213]$ , $y \in [135; 485]$
$\Omega_8$   Pipe (conf. degree= 0.52)
    $x \in [91; 286]$ , $y \in [372; 396]$
$\Omega_9$   Pipe (conf. degree= 0.47)
    $x \in [303; 304]$ , $y \in [406; 467]$
$\Omega_{10}$   Floodgate (conf. degree= 0.55)
    $x \in [293; 333]$ , $y \in [466; 485]$
$\Omega_{11}$   Floodgate (conf. degree= 0.68)
    $x \in [149; 182]$ , $y \in [375; 389]$
$\Omega_{12}$   Floodgate (conf. degree= 0.26)
    $x \in [199; 209]$ , $y \in [189; 194]$
$\Omega_{13}$   Floodgate (conf. degree= 1.00)
    $x \in [363; 369]$ , $y \in [182; 224]$
$\Omega_{14}$   Floodgate (conf. degree= 0.10)
    $x \in [302; 311]$ , $y \in [120; 130]$

The first results show that the capabilities of the system seem to be interesting. But, a problem appears with the global confidence degree : the matching quality decreases as the description is more detailed. The 3 examples given explain very well this aspect.

**- A Priori Description N.1 -
red floodgate**

**Result** :
hypothesis 1   $\pi = 1.00$
$\Omega_{13}$        red floodgate: $\pi = 1.00$
hypothesis 2   $\pi = 0.68$
$\Omega_{11}$        red floodgate: $\pi = 0.68$
hypothesis 3   $\pi = 0.10$
$\Omega_{14}$        red floodgate: $\pi = 0.10$

**- A Priori Description N.2 -
horizontal pipe *on* red floodgate**

**Result** :
hypothesis 1   $\pi = 0.68$
$\Omega_5$          horizontal pipe: $\pi = 0.76$
$\Omega_{11}$        red floodgate: $\pi = 0.68$
$R(\Omega_5, \Omega_{11})$   on: $\pi = 1.00$
hypothesis 2   $\pi = 0.10$
$\Omega_2$          horizontal pipe: $\pi = 0.88$
$\Omega_{14}$        red floodgate: $\pi = 0.10$
$R(\Omega_2, \Omega_{14})$   on: $\pi = 1.00$

**- A Priori Description N.3 -
vertical elongated pipe *elbow* horizontal pipe
*on* red floodgate**

**Result** :
hypothesis 1   $\pi = 0.10$
$\Omega_7$          vertical elongated pipe: $\pi = 0.74$
$\Omega_2$          horizontal pipe: $\pi = 0.88$
$\Omega_{14}$        red floodgate: $\pi = 0.10$

R($\Omega_7,\Omega_2$)     elbow: $\pi = 1.00$
R($\Omega_2,\Omega_{14}$)    on: $\pi = 1.00$

When the number of objects and relations increase, the *min* is a too strong mode of combination of the confidence degrees (example of the 3 Descriptions). It is contradictory with the the fact that only one hypothesis is possible for the Description N.3 ! It seems important to bring a reinforcement that can be explained like the inference of the recognition for the part of the way that is the less recognized. Classic approaches [DP88, DP92] introduce so-called *reinforcement rules*. In a first time, we had worked on a combination mode like $\sqrt[n]{\pi_1 \cdot \pi_2 \cdot ... \cdot \pi_n}$ because this rule has good properties of reinforcement, regularity... The problem concerns not only the choice of a "good" rule but its justification too...

### 3.2 Discussion

The general problem at hand is to find an object by matching a description provided by an human operator with information delivered by sensors. We assume that a description is simply a set of items that are supposed to be present in the environment. Indeed, there are several reasons why the information cannot match perfectly the description :

_ some description elements might be erroneous because the operator does not know exactly the state of the world,

_ some objects could be hidden to the sensors, so they do not appear in the information,

_ sensors might report false information because, for instance, of shadows that are interpreted as real objects,

_ sensor data processing usually recognize individual objects with limited confidence *(case of the examples given)*.

These troubles are really common and well known, and this is the reason why operators usually provide more details than strictly necessary: as it is quite natural that some of them fail to be matched, redundancy is meant for allowing the object at hand to be identified.

Unfortunately, with classic likelihood aggregation rules, because of this imperfect individual details matching, likelihood cannot increase as new information is matched with the given description; even, it usually decreases. This is contradictory with human behavior as, for instance, when 10 details are given for finding an object, one considers he has probably found the good one when only 8 are fairly found. Classic approaches [DP88, DP92] introduce so-called *reinforcement rules* without indicating why they could be allowed.

The section 4 aims at formally defining redundancy so as to derive less pessimistic likelihood aggregation rules, better fitted to the problem at hand than the classic ones.

## 4 Redundant descriptions matching

### 4.1 Defining common sense redundancy

#### 4.1.1 Preliminary definitions and hypothesis

We define the *operator provided description* as a set $D$ of *items*. Similarly, we define the *sensor delivered information* as a set $I$ of *percepts*. Any subset $D_n \subset D$ will be called *description-subsets* (*SubDs* for short), and any subset $I_m \subset I$ will be called *information-subsets* (*SubIs* for short). It can be noticed that there exist $2^{|I|}$ different sub-information.

We assume that there exists a domain-dependent matching algorithm that evaluate any pair $M_n^m = (D_n, I_m)$ so as to elaborate a *matching performance*. Typically, it is a pair *(likelihood, non-ambiguity)*.

Let us assume that a performance threshold is required for accepting a pair such as $M_n^m$. Then, let us define *maximal SubDs* as SubDs that fulfill this performance requirement and such that if we add any description item, no SubI can provide acceptable matching performance. It is important to notice that this performance requirement includes two aspects : a minimum likelihood threshold and a minimum non-ambiguity threshold.

The likelihood measures how much $D_n$ and $I_m$ are compatible. It induces an order relationship among the SubIs with a *maximal SubI* (maximal = the one that gives the maximum likelihood with $D_n$). There might be several such maximal SubIs.

The non-ambiguity indicates how much other SubIs cannot be good candidates. For instance, it could be measured by the difference between the likelihood of the maximal SubI and the one of its best competitor. At last, let us define a A *matching candidate* as a pair $M_n^m = (D_n, I_m)$ where $D_n$ is a maximal SubD (given a performance threshold), and $I_m$ is a maximal SubI (given $D_n$).

#### 4.1.2 A principle of matching using redundancy

Using redundancy to overcome the possible erroneous items issue is equivalent to declare that the exact description of the expected environment is a subpart of the description (referred to as *sub-description*).

The human user does not intentionally introduce false items in the description : he has a cooperative behavior. However, he does not provides an exhaustive description of the real scene : he selects the items that are essential to recognize the place of interest. So the sensor provided information contains percepts concerning real objects that have been discarded by the human user when choosing the description. Furthermore, the information contains unavoidable artifacts.

Consequently, an appropriate matching process will try to match the largest sub-description, but will not consider how much of the sensor provided information is not used. This is quite simple and intuitive, but a difficulty raises for applying this principle. It has been mentioned that individual items are not perfectly recognized; then, there is a permanent dilemma between declaring badly matched item and discarding the item (redundancy use), i.e. reducing the sub-description. The pattern matching problem with redundant prototype is in fact a two-criterion optimization problem: minimizing the use of redundancy while maximizing the matching performance..

### 4.2 Evaluation of redundant description matching

While any matching candidate performance is delivered by the basic domain-dependent matching algorithm, the use of redundancy is a more general concept that has to be clarified.

Given a matching candidate $(D_n, I_m)$, we call *used redundancy* the number of description items that have been discarded when reducing the description from $D$ to $D_n$. In order to assess whether this contraction is abusive or not, the use of redundancy should be compared to the number $\delta$ of redundant items that the user introduced in the initial description $D$; we call it : *description redundancy.*

For instance, if the user provided the recognition system with a description consisting of 10 items, 2 of which being redundant, it means that the user found it quite likely that 2 items in the description might be not matched, either because they are erroneous or because matching percepts are difficult to sense. Consequently, as soon as the system matches correctly 8 items, mismatching the two last ones should not decrease the matching performance, unlike classic aggregation rules do.

To apply this principle, the description redundancy has to be known. Of course, the user could be required to post this information $\delta$. However it seems possible to derive it from the basic matching algorithm behavior.

Imagine new items are discarded from $D_n$. Then, the basic matching algorithm will output better likelihood but worse non-ambiguity. There is a limit $k_n \subset D_n$ such that if any item is removed from $k_n$ then the performance no longer meets the performance requirement because of the non-ambiguity worsening. $k_n$ is called the *kernel* of $D_n$.

The result is then : if the matching $(D_n, I_m)$ is correct, then the description redundancy is $\delta$:

$$\delta = |D| - |k_n|$$

Notice that there might exist several kernels, in which case it is prudent to retain the larger one.

### 4.3 Computing matching performance when redundancy

Assume an initial description $D$ and an initial information $I$. Consider a matching candidate $(D_n, I_m)$. The preceding considerations shown that, given $D_n$, the $D$ description redundancy $\delta$ can be calculated thanks to the basic matching algorithm.

Classically, the basic matching algorithm evaluates the matching performance of $(D_n, I_m)$ by aggregating individual $D_n$ items matching performance. The matching algorithm is domain-dependent, but reasonable ones are such that the more the number of items in $D_n$, the least likelihood and the least ambiguity too.

Then, we propose to define the *matching performance with redundancy* as :

$$\hat{P}(D_n) = (\text{likelihood}(k_n), \text{ambiguity}(D_n))$$

In other words, we consider the likelihood of the minimal non-ambiguous subset of $D_n$, while for the ambiguity we take benefits of the maximum score of items matched.

For instance, assume that $D$ contains 10 items, and that we find a matching candidate $(D_n, I_m)$ where $D_n$ contains 8 items. Assume that we find out that only 6 items from $D_n$ are strictly necessary for the non-non-ambiguity. Then, when computing the likelihood we take into account the best 6 individual matchings among the 8 ones that $D_n$ implies. For assessing the non-ambiguity, we look for the best SubI $I'$ different from $I_m$ for matching $D_n$ and we compare the likelihood of $D_n, I'$ to the one of $(D_n, I_m)$.

### 4.4 Example

The problem is to find a pipe with the description:

(pipe(p), horizontal(h), long(l), blue(b))

Assume that the perception system discovers 3 regions (R1,R2,R3) that can be the representation of pipes for which the conformity degrees can be evaluated:

|    | p   | h   | l   | b   |
|----|-----|-----|-----|-----|
| R1 | 1.0 | 0.8 | 0.4 | 0.1 |
| R2 | 1.0 | 0.2 | 0.8 | 0.1 |
| R3 | 1.0 | 0.9 | 0.7 | 0.5 |

The conformity degrees are interpreted as possibilities. For declaring the pipe found, the matching possibility should be at least 0.6 and the non-ambiguity at least 0.3 . If the matching has to consider all the attributes:
        **p AND h AND l AND b**
then none of the 3 regions gives a matching better than 0.5 .

The idea hidden behind the proposed method is to assume that one of the areas corresponds to the wanted pipe, at that the bad results concerning the attribute

"b" should be the consequence of some sensing problem. That could be formalized in the default reasoning framework, i.e. the matching goal could be stated as:
(h OR problem(h)) AND (l OR problem(l)) AND
(bl OR problem(b))

Therefore, matching should discard the "b" attribute. For performing automatically this reasoning, we consider all attributes subsets: (h l), (l b) , (b h), then (h) (l) (b).

Only (h l) meets the two SubD definition conditions:
_ meets the performance requirements since with SubI R3 its possibility is 0.7 and its non-ambiguity is 0.3 (R3 best competitor for (h l) is R1 that correspond to a possibility of 0.4) ;
_ if another attribute (b) is added, then (h l b) does not meet the performance requirements.

Then, in order to assess the description redundancy, we try to discard other attributes from the maximal SubD (h l). The attribute "h" should not be dropped, otherwise R2 would rise as the best explanation for the remaining "l". However, the attribute "l" may be dropped as R3 is the best explanation for "h", with a possibility equal to 0.9 . (h) is the description kernel for the attributes (the description kernel is (ph) ).

The conclusion is then :

- the region that best matches the pipe is R3
- the description had 2 redundant elements "l" & "b"
- the matching possibility is 0.9 (= R3/h= R3/ph)
- the matching non-ambiguity is 0.4 (= R3/phlb - R2/phlb)

## Conclusion

The problem addressed in this paper deals with object recognition driven by an operator provided object-in-the-scene description. In such cases, a major difficulty is to deal with uncertainty; uncertainty is linked to perception limits, and also to possible errors in the given description.

This paper presented first a method for propagating uncertainty from initial data to matching candidates for descriptions that indicate individual objects attributes as well as n-ary objects spatial relationships. The proposed approach has been tested only with possibilities that is the framework chosen. The results with the first version of the system are presented. A particular problem appears with the global confidence degree : the matching quality decreases as the description is more detailed.

Unfortunately, classic uncertainty measurement frameworks lead to monotonic non-increasing likelihood as the description size increases, and this prevent us to use description redundancy in order to enhance the matching result. Existing reinforcement rules do not rely on convincing justifications. In this paper, a formal definition of redundancy allowed the matching likelihood to be evaluated by selecting a limited number of description items, and a method has been proposed to assess how many items, and which ones, may be dropped. The second version of the system will use these last developments.

The example of section 4 shows that, as intended, the method proposed leads to higher matching performance in the case of redundancy. But also, of course, the risk raises to recognize the object in a wrong place when classic approaches would have refused to match the description with the scene as perceived. This sort of dilemma is inevitable when dealing with decision problems.

## Acknowledgments

The authors would like to thank the STSN - DA/S/ES department of the French Defence Ministry for its support and Olivier Le Moigne for its useful comments.

## References

[CK92]   Chellapa R., Kashyap R.L., "Image Understanding", *Encyclopedia of artificial intelligence*, Wiley-Interscience : 641-663, 1992.

[DCB89]  Draper B. A., Collins R. T., Brolio J., Hanson A., Riseman E. M., *"The schema system"*, Int. Journal of Computer Vision, 2;209-250, 1989.

[DP88]   Dubois D., Prade H., *Possibility Theory, an approach to the computerized processing of uncertainty*, Plenum Press, New-York, 1988.

[DP92]   Dubois D., Prade H., "Combination of information in the framework of possibility theory", *Data Fusion in Robotics and Machine Intelligence*, Academic Press, 1992.

[Jau86]  Jaulent M.C., *A system for identifying geometrical plan objects described by a human operator*, Ph.D. Thesis, I.N.P. Toulouse, 1986. (in French)

[Lem95]  Lemaire J., *Development of a scene recognition system with imprecise descriptions*, report DERA 200/95, 1995.(in French)

[Lem96]  Lemaire J.,"Use of A Priori Description in High Level Language and Management of the Uncertainty in a Scene Recognition System", *Proceedings of 13th. ICPR*, Vienna, August 1996.

[LLB96]  Lemaire J., Le Moigne O., Barrouil C., "How to manage uncertainty in scene recognition", *Proceedings of IPMU'96*, Granada, July 1996.

[SG94]   Sandakly F., Giraudon G., "Multispecialist system for 3D scene analysis", *Proceedings 11th. ECAI* : 771-775, August 1994.